\documentclass[lettersize,journal]{IEEEtran}
\usepackage{amsmath,amsfonts}
\usepackage{algorithmic}
\usepackage{algorithm}
\usepackage{array}
\usepackage[caption=false,font=normalsize,labelfont=sf,textfont=sf]{subfig}
\usepackage{textcomp}
\usepackage{stfloats}
\usepackage{url}
\usepackage{verbatim}
\usepackage{graphicx}
\usepackage{cite}
\usepackage{multirow}
\usepackage{threeparttable}
\usepackage{tikz}
\usepackage[edges]{forest}
\usepackage{bbm}
\usepackage{pbox}
\usepackage{hyperref}

\usepackage{amssymb}
\usepackage{booktabs}
\usepackage{bm}
\usepackage{colortbl}
\usepackage{times}
\usepackage{epsfig}
\usepackage{amssymb}
\usepackage{multirow}
\usepackage{threeparttable}
\usepackage{caption}
\usepackage{soul}
\usepackage{xcolor}
\usepackage[numbers]{natbib}

\usepackage[capitalize]{cleveref}
\crefname{section}{Sec.}{Secs.}
\Crefname{section}{Section}{Sections}
\Crefname{table}{Table}{Tables}
\crefname{table}{Tab.}{Tabs.}
\crefname{equation}{Eq.}{Eqs.}

\usepackage{silence}
\usepackage{pifont}%

\definecolor{leafcolor}{rgb}{0.8,0.8,0.8}

\begin{document}

\title{GenDet: Painting Colored Bounding Boxes on Images via Diffusion Model for Object Detection}

\author{Chen Min, Chengyang Li, Fanjie Kong, Qi Zhu, Dawei Zhao, Liang Xiao
}

\markboth{}%
{Shell \MakeLowercase{\textit{et al.}}: A Sample Article Using IEEEtran.cls for IEEE Journals}

\maketitle

\begin{abstract}
This paper presents GenDet, a novel framework that redefines object detection as an image generation task. In contrast to traditional approaches, GenDet adopts a pioneering approach by leveraging generative modeling: it conditions on the input image and directly generates bounding boxes with semantic annotations in the original image space. GenDet establishes a conditional generation architecture built upon the large-scale pre-trained Stable Diffusion model, formulating the detection task as semantic constraints within the latent space. It enables precise control over bounding box positions and category attributes, while preserving the flexibility of the generative model. This novel methodology effectively bridges the gap between generative models and discriminative tasks, providing a fresh perspective for constructing unified visual understanding systems. Systematic experiments demonstrate that GenDet achieves competitive accuracy compared to discriminative detectors, while retaining the flexibility characteristic of generative methods.
\end{abstract}

\begin{IEEEkeywords}
Object Detection, Diffusion Model, Generative Model, Probabilistic Model. 
\end{IEEEkeywords}

\section{Introduction}

\begin{figure}[t]
	\centering
	\includegraphics[width=0.6\columnwidth]{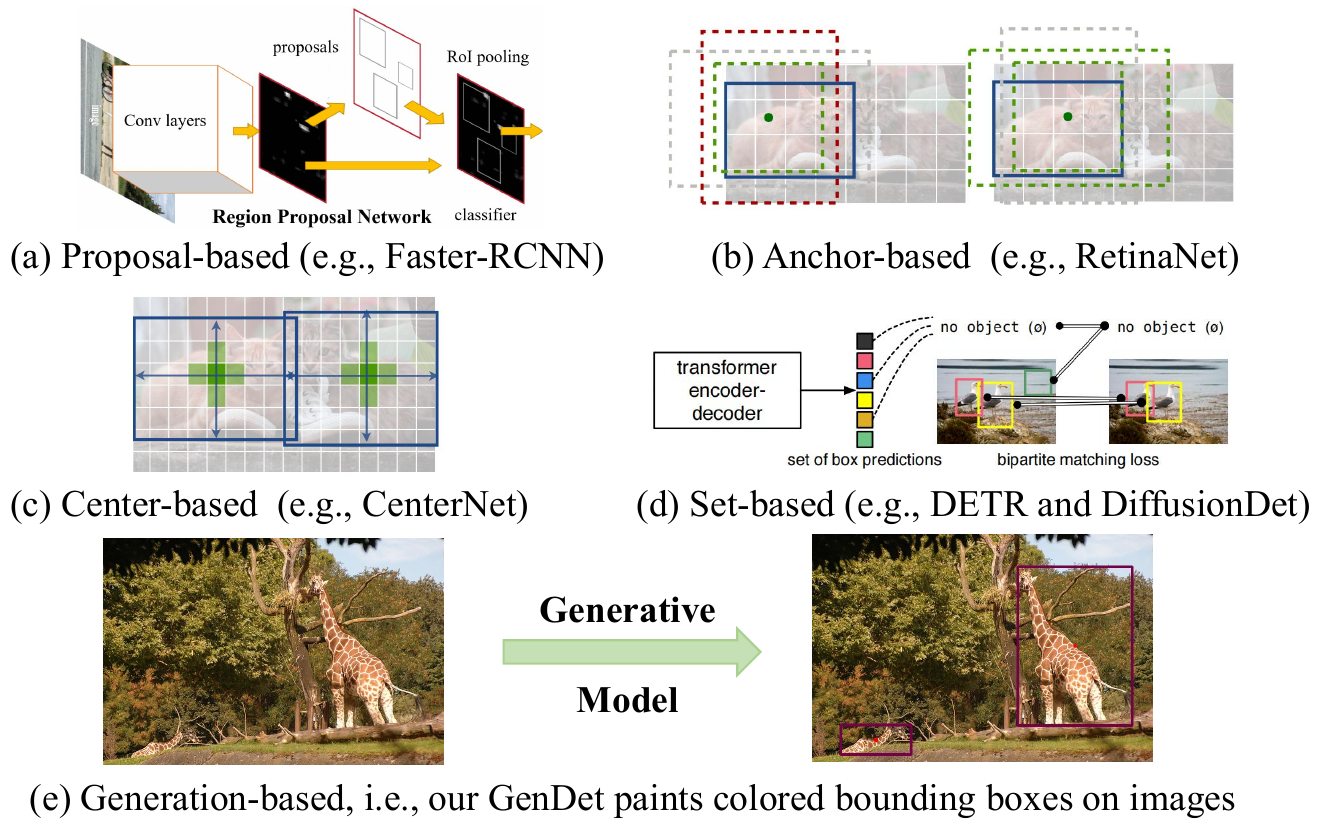}
	\caption{Difference between object detection methods.  Existing object detectors (\textit{i.e.}, (a), (b), (c), and (d)) are discriminative methods, involving the classification of object categories and the regression of bounding box dimensions. In contrast, our GenDet takes an entirely different approach by formulating object detection as image generation task, directly rendering colored object bounding boxes onto the input image. This innovative strategy offers a more direct and intuitive way to perform object detection.}
	\label{first}
\end{figure}

As a fundamental task in computer vision, object detection has evolved through various discriminative paradigms, including region proposal methods (\textit{e.g.}, Faster R-CNN~\citep{ren2016faster}), anchor box mechanisms (\textit{e.g.}, YOLO~\citep{yolo} and SSD~\citep{ssd}), center prediction approaches (\textit{e.g.}, CenterNet~\citep{centernet}), and set-based prediction techniques (\textit{e.g.}, DETR~\citep{detr}), as illustrated in Figure~\ref{first}.
While these detectors have made significant strides, they still treat object detection as a discriminative task, obtaining detection results through classification and regression~\citep{zhang2020bridging,yang2019reppoints}.

At the same time, generative models are driving a paradigm shift in visual representation learning. From latent space reconstruction in VAEs~\citep{vae} to adversarial training in GANs~\citep{gan}, and more recently, diffusion models~\citep{ddpm} that generate images through gradual denoising, especially Stable Diffusion~\citep{ldm}, which enables multimodal controllable generation, these models exhibit a powerful capacity to model visual data distributions. Generative models have already demonstrated cross-task transfer potential in applications such as image inpainting and super-resolution. However, their full potential in object detection remains underexplored. This distinction between generative and discriminative paradigms raises an important scientific question: \textit{Is it possible to develop a unified framework that enables a single generative model to perform both image synthesis and object detection?}

Current advancements~\citep{zhao2025diception} in Stable Diffusion have primarily focused on dense prediction tasks~\citep{ravishankar2025scaling} like depth estimation~\citep{marigold, depthcrafter}, surface normal prediction~\citep{dmp}, semantic segmentation~\citep{lai2023denoising}, optical flow estimation~\citep{hai2025hierarchical}, and object keypoint localization~\citep{wang2025generalizable}, while neglecting its application potential in fundamental tasks like object detection. This paper finds that Stable Diffusion, during pretraining~\citep{5b}, implicitly learns rich knowledge of object structure and spatial relationships~\citep{chen2025,dae}, which can be explicitly leveraged to enhance detection capabilities with appropriate conditioning. DiffusionDet~\citep{diffusiondet}, a model that applies diffusion models for object detection, follows the traditional discriminative framework by using diffusion models to denoise bounding boxes.
\begin{figure*}[t]
	\centering
	\includegraphics[width=0.95\linewidth]{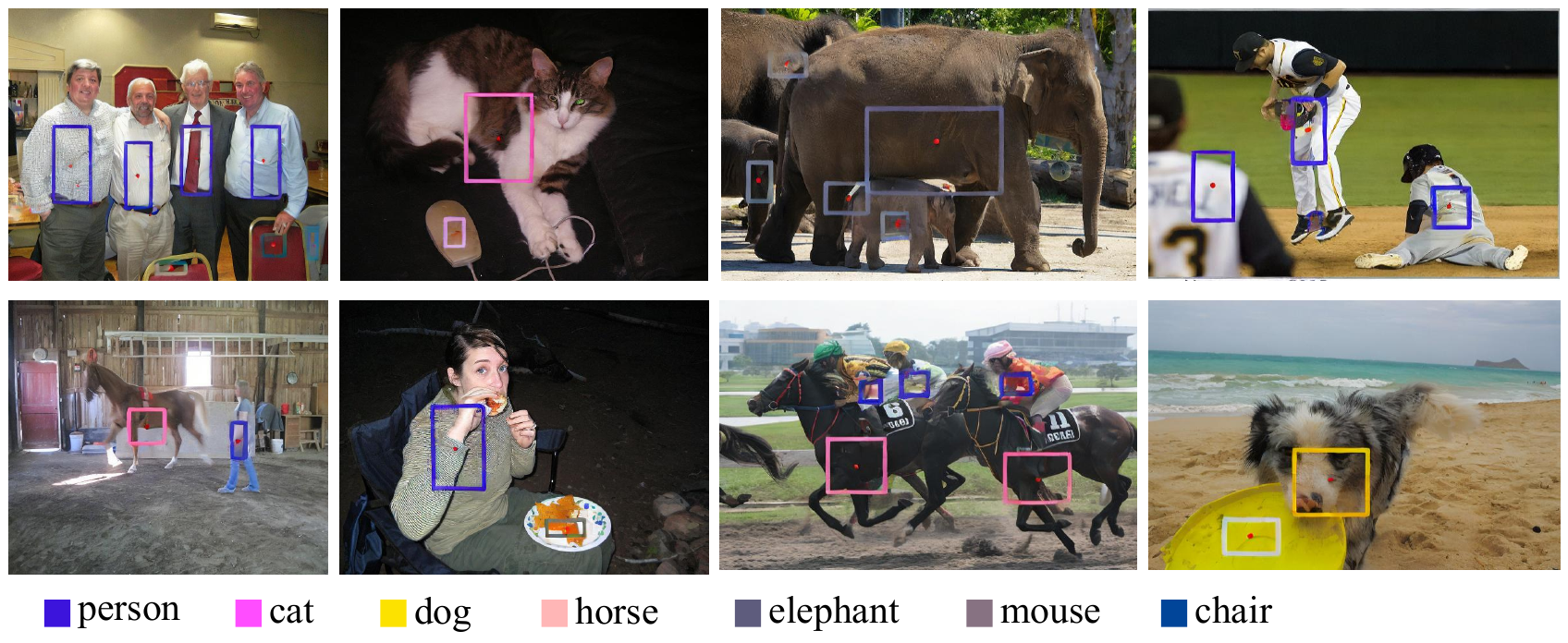}
	\caption{Visualization of the generated object bounding boxes. GenDet is a diffusion-based image generation model designed for object detection. It utilizes the prior knowledge from Stable Diffusion to directly paint detection boxes on the original image, with different colors indicating different object categories. To minimize box overlap, the bounding boxes are scaled down, and red dots are added at the center of each object to enhance recognition accuracy.}
	\label{fig:overall}
\end{figure*}

Building on the preceding analysis, we introduce GenDet, the first framework to reconceptualize object detection as a conditional image generation task. Unlike traditional detectors that rely on explicit region proposals or classification heads, GenDet conditions on the input image to model a joint distribution over object locations and semantic categories within the latent space of a generative model. Specifically, it integrates detection objectives directly into the diffusion process, allowing the model to maintain the expressive generative capabilities of the underlying architecture while producing detection-aware outputs. 

To evaluate the performance of GenDet, we conducted experiments on the COCO 2017~\citep{coco} and CrowdHuman~\citep{shao2018crowdhuman} datasets. The results in Figure~\ref{fig:overall} demonstrate that GenDet can effectively detect objects, predict bounding box dimensions, and achieve performance comparable to state-of-the-art algorithms. GenDet represents a novel and intuitive method for object detection, paving the way for future exploration of generative methodologies.

The main contributions of this work are listed below:

\begin{itemize}
	\item We propose GenDet, a novel framework that reconceptualizes object detection as a conditional image generation task. GenDet models a joint distribution over object locations and semantic categories within the latent space of a generative diffusion model, providing a unified approach to detection.
	
	\item We leverage the implicit knowledge of object structure and spatial relationships learned by pre-trained diffusion models. GenDet maintains the rich representation power of generative models while achieving competitive performance on standard object detection benchmarks, establishing a new paradigm for generative-based detection methods.
	
	\item We conduct extensive experiments on benchmarks, where GenDet demonstrates competitive performance.
\end{itemize}

\section{Related Work}

\subsection{Object Detection}

Object detection is a crucial task in computer vision, and a variety of algorithms have been proposed to address this challenge. Early object detection methods focus on classifying and regressing candidate boxes, which can be broadly categorized into proposal-based and anchor-based approaches. Proposal-based methods, such as Faster R-CNN~\citep{ren2016faster}, offered high accuracy but are computationally expensive due to their two-step design. In contrast, anchor-based methods like YOLO~\citep{yolo} and SSD~\citep{ssd} are designed for fast, real-time detection, but may struggle with accuracy, particularly for small or complex objects. CenterNet~\citep{centernet}, a center-based detection algorithm, introduced a different paradigm by directly predicting object center points and dimensions, bypassing traditional bounding box regression and eliminating the need for non-maximum suppression (NMS). Set-based methods, such as DETR~\citep{detr}, leverage transformers for end-to-end optimization and formulate object detection as a matching problem between predicted and ground-truth bounding boxes. Recently, RT-DETR~\citep{rtdetr} introduced the first real-time end-to-end object detector.
Different from the above discriminative methods, we propose a novel generation-based object detection method. Our approach directly renders colored bounding boxes on the image, offering a fresh perspective on bridging generative modeling and object detection.

\subsection{Diffusion Models} 
Denoising Diffusion Probabilistic Models (DDPMs)~\citep{ddpm} have emerged as a powerful framework for generative modeling. These models learn to reverse a diffusion process that gradually corrupts images with Gaussian noise, enabling the generation of new samples by starting with random noise and iteratively denoising. Building on the success of DDPMs, DDIMs~\citep{ddim} introduced a more efficient, non-Markovian reverse diffusion process, improving the practicality of these models for real-world applications.  
The remarkable generative capabilities of DDPMs~\citep{ddpm}~\citep{ddim}, DDIMs, and Latent Diffusion Model (LDM)~\citep{ldm} have spurred the development of conditional generation. 
For instance, Stable Diffusion~\citep{ldm} has redefined text-to-image generation by training on large-scale datasets such as LAION-5B~\citep{5b}, achieving unprecedented image synthesis quality. A key innovation of Stable Diffusion is the Latent Diffusion Model (LDM), which operates in a compressed latent space, significantly reducing computational complexity while preserving output quality.   
Building on LDMs, ControlNet~\citep{controlnet} introduced controllable generation methods, such as semantic map guidance, further expanding the applications of diffusion models to more structured and interpretable outputs.
Diffusion-based generative models have proven to be exceptionally effective in producing high-quality outputs. Building on this success, recent approaches have explored utilizing the prior knowledge embedded in Stable Diffusion, trained on large-scale datasets, for dense perception tasks such as depth estimation~\citep{diffusiondepth,depthfm,zhang2024betterdepth}, normal estimation~\citep{dmp}, and semantic segmentation~\citep{lai2023denoising}.  
For example, Marigold~\citep{marigold} and GeoWizard~\citep{geowizard} directly applied the standard diffusion framework along with pre-trained parameters to these tasks. However, their methods overlook the inherent differences between image generation and dense prediction, leading to suboptimal results. GenPercept~\citep{genpercept} and StableNormal~\citep{stablenormal} introduced single-step diffusion strategies to minimize unnecessary variations and improve the consistency of predictions. 
%Lotus~\citep{lotus} took a more comprehensive approach by systematically analyzing the stochastic diffusion formulation for dense prediction tasks. 
\citep{ravishankar2025scaling} unify tasks such as depth estimation, optical flow, and amodal segmentation under the framework of diffusion model, but overlook object detection—a fundamental visual perception task.
Motivated by these advancements, we extend the capabilities of Stable Diffusion to object detection tasks, leveraging their prior knowledge learned from large-scale data. 
\begin{figure*}[t]
	\centering
	\includegraphics[width=0.99\linewidth]{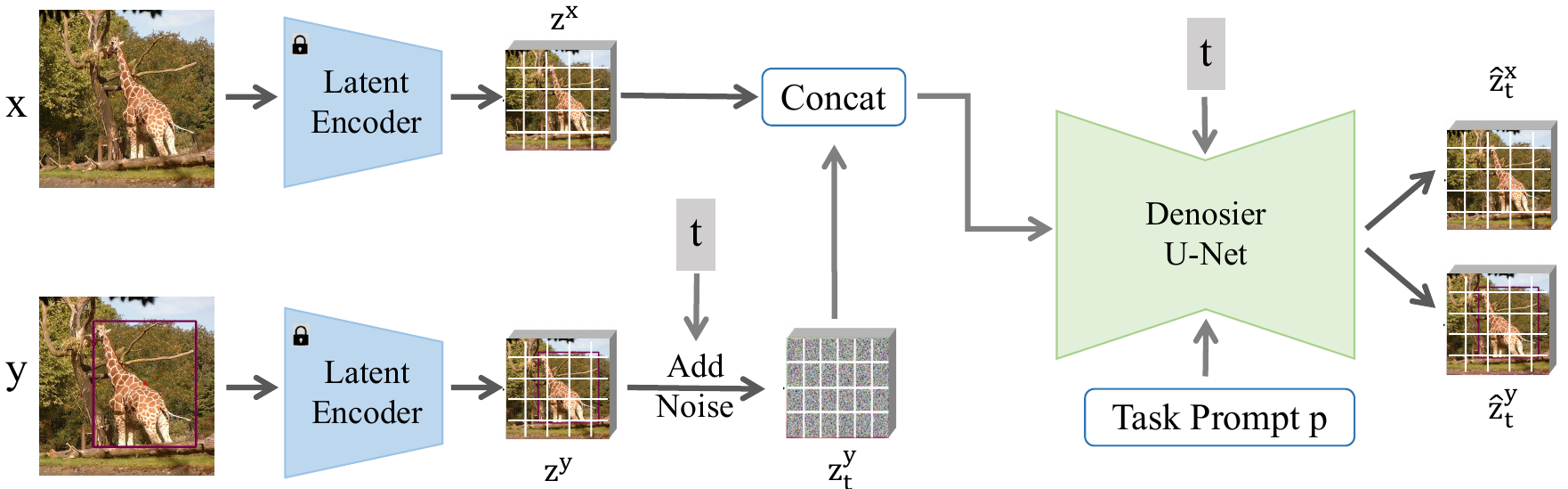}
	\caption{Overview of GenDet's training pipeline. Starting from a pre-trained Stable Diffusion model, both the image \(x\) and its corresponding annotation image \(y\) with colored bounding boxes are encoded through the pre-trained VAE. The noisy version of the annotation image, \(z_{t}^y\), is obtained by introducing noise at a specific diffusion step \(t \in [1, T]\). The U-Net~\citep{unet} input layer is modified to process the concatenated inputs, and the model is then fine-tuned with the standard diffusion objective, \(v\)-prediction~\citep{v}, following the multi-step training procedure. Additionally, the task prompt p is introduced to either generate annotation image \(y\) or reconstruct the input image \(x\).}
	\label{fig:flowchart}
\end{figure*}
\subsection{Diffusion Models for Object Detection} 
Diffusion models have found applications in object detection, generally falling into two main categories. One approach leverages diffusion models for data augmentation~\citep{fang2024data}, and the other focuses on denoising bounding boxes~\citep{diffusiondet}. DiffusionEngine~\citep{zhang2023diffusionengine} enhances the scalability and diversity of high-quality detection training pairs. \citep{li2024simple} propose a data augmentation framework for object detection using text-to-image models. ODGEN~\citep{zhu2025odgen} enables controllable image generation using bounding boxes and text prompts, allowing the creation of high-quality data for complex scenes.
In addition to data generation, some methods focus on the denoising process for bounding box refinement. DiffusionDet~\citep{diffusiondet} treats object detection as a denoising diffusion task, refining noisy bounding box predictions into accurate ones. Similarly, Diffusion-SS3D~\citep{ho2023diffusion} introduces noise to simulate corrupted 3D object sizes and class labels, applying the diffusion model to denoise and recover bounding box outputs. MonoDiff~\citep{ranasinghe2024monodiff} and CoDiff~\citep{huang2025codiff} utilize the reverse diffusion process to estimate 3D bounding boxes.
Different from these methods, we propose GenDet, which directly generates object bounding boxes on the original image to perform object detection tasks.

\section{Methodology}

Unlike existing object detection methods, which approach detection as a discriminative task, we treat object detection as a task of image-conditioned annotation generation. Our goal is to leverage the pre-trained Stable Diffusion~\citep{ldm} as a prior for the object detection task, enabling the learning of the conditional distribution \( D(\mathbf{y}|\mathbf{x}) \) using a labeled training dataset \( D = \{(\mathbf{x}_i, \mathbf{y}_i)\}_{i=1}^n \), where \( \mathbf{x}_i \) is the input image and \( \mathbf{y}_i \) represents the corresponding annotation image with colored bounding boxes. Both \( \mathbf{x}_i \) and \( \mathbf{y}_i \) are elements of \( \mathbb{R}^{H \times W \times 3} \). In Section~\ref{pre}, we first introduce the Stable Diffusion used as the prior. Then, in Section~\ref{odm}, we describe the proposed GenDet method.

\subsection{Preliminaries}
\label{pre}

Stable Diffusion~\citep{ldm} operates in a low-dimensional latent space. The latent space is formed at the bottleneck of a variational autoencoder (VAE)~\citep{vae}, which is trained separately from the denoiser. This setup allows for efficient compression of the latent space and ensures perceptual alignment with the data space. The process begins with an autoencoder, \( \{\mathcal{E} (\cdot), \mathcal{D}(\cdot)\} \), which is trained to map between the RGB space and the latent space, such that \( \mathcal{E}(\mathbf{x}) = \mathbf{z}^\mathbf{x} \) and \( \mathcal{D}(\mathbf{z}^\mathbf{x}) \approx \mathbf{x} \). These autoencoders also map dense annotations effectively into the latent space, i.e., \( \mathcal{E}(\mathbf{y}) = \mathbf{z}^\mathbf{y} \) and \( \mathcal{D}(\mathbf{z}^\mathbf{y}) \approx \mathbf{y} \). Stable Diffusion incorporates a pair of forward noise addition and reverse denoising processes within the latent space. In the forward process, Gaussian noise is incrementally added to the sample \( \mathbf{z}^\mathbf{y} \) at each time step \( t \in [1, T] \), resulting in the noisy sample $\mathbf{x}_t^\mathbf{y}$:
\begin{equation}
	\mathbf{z}_t^\mathbf{y} = \sqrt{\bar{\alpha}_t} \mathbf{z}^\mathbf{y} + \sqrt{1 - \bar{\alpha}_t} \epsilon,
\end{equation}
where \( \epsilon \sim \mathcal{N}(0, I) \) represents the noise sampled from a standard normal distribution, and \( \bar{\alpha}_t := \prod_{s=1}^{t} (1 - \beta_s) \) defines the cumulative product of the noise schedule, with \( \{\beta_1, \beta_2, \dots, \beta_T\} \) being the set of noise coefficients over \( T \) diffusion steps. At the final time-step \( T \), the sample \( \mathbf{z}^\mathbf{y} \) becomes pure Gaussian noise. In the reverse diffusion process, a neural network \( f_\theta \), typically implemented as a U-Net~\citep{unet}, is trained to progressively remove noise from \( \mathbf{z}_t^\mathbf{y} \), reconstructing the clean sample \( \mathbf{z}^\mathbf{y} \). The training procedure involves randomly selecting a time-step \( t \in [1, T] \) and minimizing the associated loss function \( \mathcal{L}_t \).

DDIM~\citep{ddim} is a crucial method for accelerating the sampling process in multi-step diffusion models. It introduces an implicit probabilistic framework that allows for a significant reduction in the number of denoising steps, all while preserving the quality of the generated output. Specifically, the denoising operation transitioning from \( \mathbf{z}_{\tau}^\mathbf{y} \) to \( \mathbf{z}_{\tau-1}^\mathbf{y} \) is described as follows:
\begin{equation}
	\mathbf{z}_{\tau-1}^\mathbf{y} = \sqrt{\bar{\alpha}_{\tau-1}} \hat{\mathbf{z}}_{\tau}^\mathbf{y} + \text{direction}(\mathbf{z}_{\tau}^\mathbf{y}) + \sigma_\tau \epsilon_\tau.
\end{equation}
At each denoising step \( \tau \), \( \hat{\mathbf{z}}_\tau^\mathbf{y} \) denotes the model’s prediction of the clean sample, while \( \text{direction}(\mathbf{z}_{\tau}^\mathbf{y}) \) indicates the vector pointing towards \(\mathbf{z}_{\tau}^\mathbf{y} \). The term \( \sigma_\tau \) is used to control the amount of noise added, and it can be set to zero when deterministic denoising is desired. The sequence \( \tau \in \{\tau_1, \tau_2, \dots, \tau_S\} \) represents a subset of time steps selected from \( [1, T] \), allowing for efficient sampling. During inference, DDIM performs iterative denoising, progressively refining the sample from \( \tau_S \) down to \( \tau_1 \), ultimately producing a clean output.

These concepts serve as the foundation for the proposed GenDet framework, as discussed in subsequent sections.

\subsection{GenDet}
\label{odm}
Next, we first introduce the generated target image $\mathbf{y}$, which contains the object detection bounding box information. Then, we discuss techniques for reducing the randomness in the generative model. Finally, we outline the loss function, inference process, and post-processing steps.
\subsubsection{Conditional Generation Architecture}
\begin{figure}[h]
	\centering
	\includegraphics[width=0.99\linewidth]{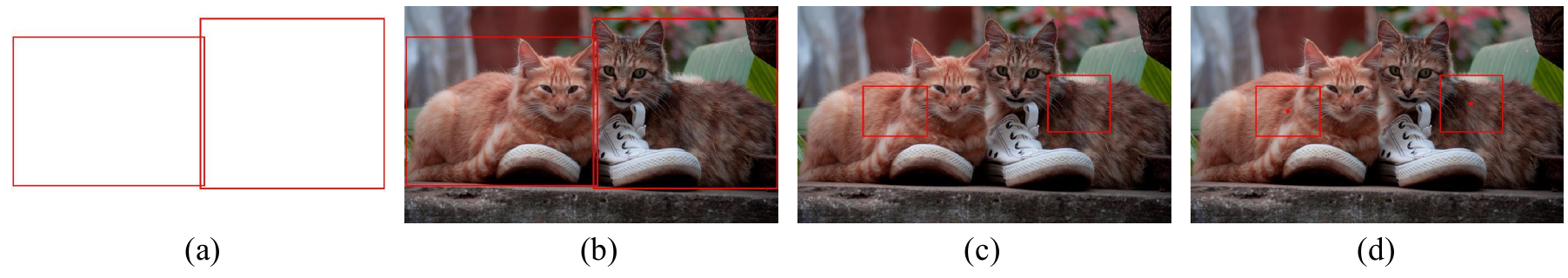}
	\caption{Illustration of different types of target images \(\mathbf{y}\), which encode object detection bounding box information. To better align with the image generation process in Stable Diffusion~\citep{ldm}, we overlay bounding boxes on the original image, using distinct colors to differentiate object categories. To reduce overlap, as shown in (d), we shrink the bounding boxes and further enhance detection cues by marking the center of each box with a red dot.}
	\label{fig:type}
\end{figure}

\begin{figure}[t]
	\centering
	\includegraphics[width=0.9\linewidth]{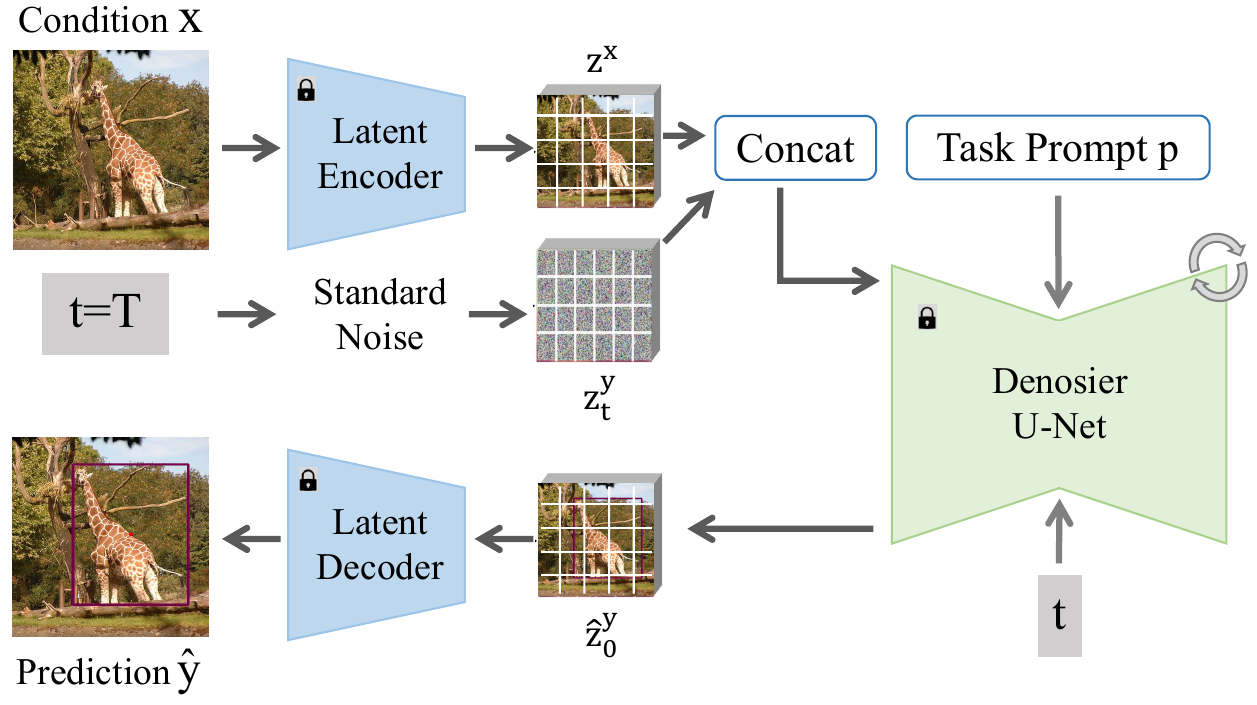}
	\caption{Overview of the GenDet inference scheme. Given an input image \(\mathbf{x}\), GenDet begins by encoding it using the pre-trained Stable Diffusion VAE to generate the latent code \(\mathbf{z}^x\). This latent code is then combined with the annotation image's latent \(\mathbf{z}_t^\mathbf{y}\) and fed into the modified, fine-tuned U-Net~\citep{unet} at each denoising iteration. After completing \(T\) steps of the diffusion process, the resulting latent \(\mathbf{z}_0^\mathbf{y}\) is decoded into the prediction image $\hat{\mathbf{y}}$. The final object detection output is obtained by applying post-processing to $\hat{\mathbf{y}}$.}
	\label{fig:infer}
\end{figure}
To enable object detection tasks with Stable Diffusion, it is crucial to transform object detection targets into a format compatible with the generative model’s output. As illustrated in Figure~\ref{fig:type}, we explored various strategies for generating object detection results using generative models.  

Our initial approach involved generating bounding boxes on a plain white background, as shown in Figure~\ref{fig:type}(a). However, since Stable Diffusion~\citep{ldm} is trained on a vast dataset of diverse images, introducing a uniform background disrupts its learned data distribution. Our experiments confirmed that this approach significantly hindered the model’s ability to generate accurate bounding boxes.  

To address this limitation, we explored an alternative strategy by preserving the input image’s distribution and directly overlaying bounding boxes on the original image, using different colors to distinguish object categories, as depicted in Figure~\ref{fig:type}(b). However, this method led to substantial overlap when multiple objects were present, making it challenging to discern individual detections.  
To mitigate this issue, we reduced the size of the bounding boxes, minimizing overlap while maintaining object visibility, as illustrated in Figure~\ref{fig:type}(c). Furthermore, to facilitate easier detection, we enhanced detection cues by marking the center of each bounding box with a red dot, as shown in Figure~\ref{fig:type}(d).  
This analysis informed the design of our prediction targets for object detection using generative models, ensuring compatibility with Stable Diffusion while optimizing detection accuracy and efficiency.

\subsubsection{Training Pipeline}

We leverage the pre-trained Stable Diffusion~\citep{ldm} framework as the foundation model, allowing us to tap into the robust and transferable image priors from LAION-5B~\citep{5b}, while efficiently learning distribution priors in a low-dimensional latent space, requiring only minimal changes to the U-Net~\citep{unet} architecture.

As illustrated in Figure~\ref{fig:flowchart}, both the image \(\mathbf{x}\) and its corresponding annotation image \(\mathbf{y}\), which includes colored bounding boxes, are first encoded using the pre-trained VAE to the latent space. To introduce controlled noise, a noisy version of the annotation image, $\mathbf{z}_t^\mathbf{y}$, is created by adding noise at a specific diffusion step \(t \in [1, T]\). The input layer of the U-Net is modified to handle the concatenation of the original image and the noisy annotation, allowing the model to process both simultaneously. The model is then fine-tuned using the standard diffusion objective, as part of a multi-step training procedure. This setup ensures that the model gradually refines its understanding by progressively denoising the annotation image, leading to more accurate predictions over time.

\subsubsection{Dual-Path Conditional Injection}
Generating target bounding boxes in Stable Diffusion can lead to inaccuracies, especially for small objects, which are difficult to reconstruct with precision. We introduce a dual-path conditional injection mechanism with task prompt~\citep{driveworld,lotus} and train the model to generate both the input image \(\mathbf{x}\) and its corresponding detection annotation \(\mathbf{y}\).

The dual-path conditional injection mechanism allows the denoising model \(f_\theta\) to dynamically switch between generating annotations \(\mathbf{y}\) and reconstructing the input image \(\mathbf{x}\). When the prompt is set to \(p_y\), the model concentrates on producing the annotation \(\mathbf{y}\), whereas, when set to \(p_x\), it reconstructs the image \(\mathbf{x}\). The task prompt \(p\) is represented as a one-dimensional vector, which is encoded using a positional encoder and integrated with the time embeddings of the diffusion model. This setup ensures a smooth transition between tasks, preventing any cross-task interference.
This approach enhances the model's ability to make more accurate predictions, ultimately improving overall performance in bounding box generation.

\subsubsection{Multi-Grained Training Objective}
To ensure that the generated images with detection boxes retain fine-grained geometric structures, we introduce a gradient loss~\citep{ma2020structure}. Let \( \mathbf{m} \) denote either \( \mathbf{x} \) or \( \mathbf{y} \) depending on the task prompt \( p \). This loss encourages the preservation of geometric details, resulting in more photo-realistic outputs. Specifically, the gradient loss is formulated by minimizing the distance between the gradient map extracted from the generated image and that from the corresponding ground truth image.
\begin{equation}
	\mathcal{L}_{t}^{gra} = \alpha_t\mathbb{E}_{\hat{z}^m_t, z^m_t} \left\| G(\hat{z}^m_t) - G(z^m_t)) \right\|_1,
\end{equation}
where $\alpha_t$ is a weighting factor inversely proportional to the timestep $t$, placing higher emphasis on later denoising steps closer to the original image. The gradient magnitude is defined as \( G(z) = \| \nabla z \|_2^2 \), which is employed in the gradient loss to encourage the preservation of fine-grained geometric details. The squared \( L_2 \) norm \( \| \nabla z \|_2^2 \) quantifies the overall intensity of the image gradients, promoting sharper object boundaries and more consistent structural reconstruction.
The gradient vector at position \((u, v)\) is calculated as:
\begin{equation}
	\begin{aligned}
		z_u(u, v) &= z(u + 1, v) - z(u - 1, v), \\
		z_v(u, v) &= z(u, v + 1) - z(u, v - 1), \\
		\nabla z(u, v) &= \left( z_u(u, v), z_v(u, v) \right).
	\end{aligned}
\end{equation}

The overall loss function for conditional image generation, shown below, facilitates the simultaneous learning of both image appearance and object boundaries:
\begin{equation}
	\mathcal{L}_t^{all} = \lambda_1\mathbb{E}_{\mathbf{x},\mathbf{m},\epsilon ,t,p} [  \|\ f_\theta(\mathbf{z}_t^\mathbf{m}; \mathbf{x}, p) - \mathbf{\hat{z}}_t^\mathbf{m}\|_2^2  + \lambda_2\mathcal{L}_{t}^{gra},
\end{equation}
where $\lambda_1$ and $\lambda_2$ are weighting coefficients that balance the contributions of pixel-level reconstruction and gradient-based geometric preservation.

\subsubsection{Inference Pipeline}

The overall inference pipeline is shown in Figure~\ref{fig:infer}. First, we encode the input image \(x\) into the latent space \(\mathbf{z}^x\) using the pre-trained Stable Diffusion VAE, while initializing the annotation image latent as Gaussian noise. This latent is then iteratively denoised according to the same schedule used during fine-tuning. For faster inference, we utilize DDIM’s~\citep{ddim} non-Markovian sampling strategy with re-spaced steps. The resulting latent code $\hat{\mathbf{z}}_0^\mathbf{y}$ is decoded into the final prediction image \( \hat{\mathbf{y}} \) using the VAE decoder, followed by post-processing to obtain the final object detection results.

\subsubsection{Feature-based Post Processing}
\begin{figure}[t]
	\centering
	\includegraphics[width=0.99\linewidth]{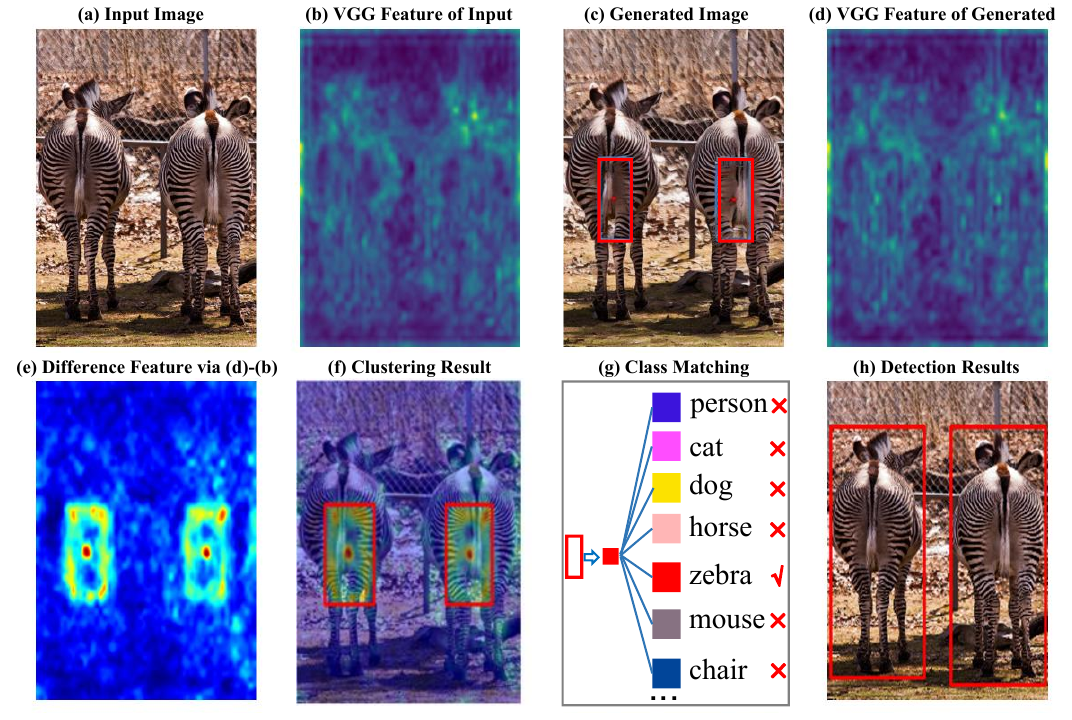}
	\caption{Illustration of feature-based post processing for generation-based object detection.}
	\label{fig:post}
\end{figure}
After the training process outlined above, GenDet is capable of generating images with colored bounding boxes, as shown in Figure~\ref{fig:overall}. To determine the final object detection box size and category, we propose feature-based post-processing method.

%\paragraph{Rule-based Post Processing.}

We observe that although the images Figure~\ref{fig:post}(c) generated by the diffusion model are not pixel-wise identical to the input images Figure~\ref{fig:post}(a), they exhibit strong similarity in high-dimensional convolutional network features. As illustrated in Figure~\ref{fig:post}(b) and (d), we first extract VGG16~\citep{vgg} features from both the original image and the generated image (with colored bounding boxes). By computing the difference between these feature representations, we obtain a feature difference map, as shown in Figure~\ref{fig:post}(e). By removing pixels with negligible differences, we further refine this to obtain the map in Figure~\ref{fig:post}(f). Subsequently, clustering methods such as DBSCAN~\citep{ester1996density} are applied to group the remaining significant pixels, resulting in the localization of object bounding boxes. We then extract the color of the pixels around each bounding box in Figure~\ref{fig:post}(g) from the image in Figure~\ref{fig:post}(c), and match this color to a set of predefined ground-truth color classes, which encode semantic object categories and spatial locations.

Through this matching process, we can infer both the object class and the corresponding bounding box dimensions, effectively recovering a structured representation from the visual artifacts left by the generative process.
However, this feature-based approach encounters limitations in complex scenarios, particularly in crowded scenes, or when dealing with occluded and small-scale objects. In such cases, the visual cues may become ambiguous or indistinct, making it difficult for heuristic methods to accurately distinguish individual objects. 
\subsubsection{Learning-based Post Processing}
\begin{figure}[t]
	\centering
	\includegraphics[width=0.9\linewidth]{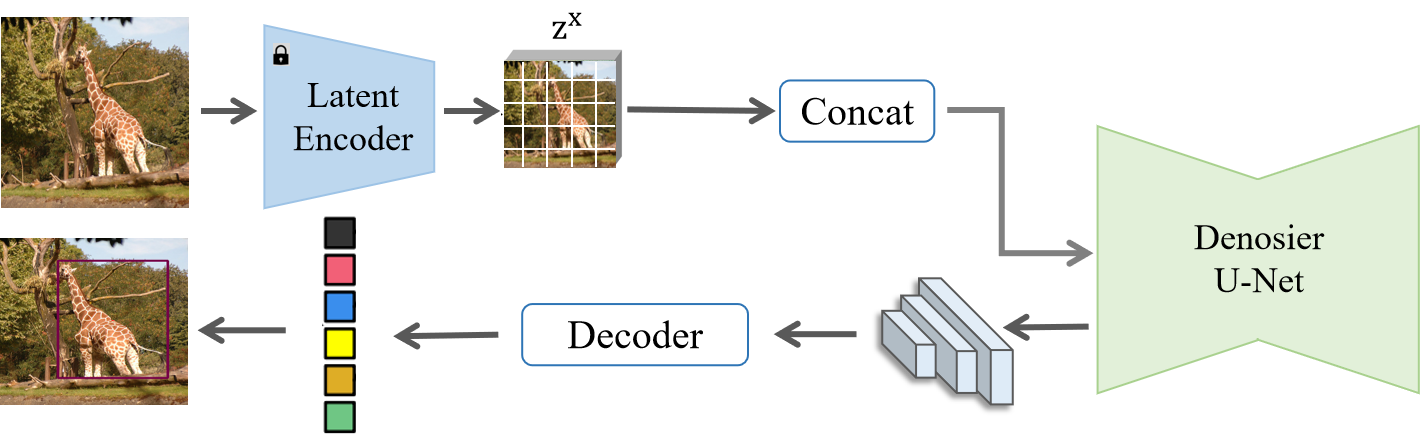}
	\caption{Illustration of learning-based post-processing for object detection with multi-scale diffusion features.}
	\label{fig:post2}
\end{figure}
To address the limitations of feature-based methods in complex visual scenarios, we propose a learning-based approach that incorporates a set-based object detection head. This head leverages the rich prior features embedded in the diffusion model, which generates colored bounding boxes corresponding to different object classes.
As illustrated in Figure~\ref{fig:post2}, we specifically utilize the previously trained diffusion model capable of generating colored bounding boxes. The model captures rich semantic information across multiple scales, which serves as a strong prior. Similar to RT-DETR~\citep{rtdetr}, a set-based object detector head is trained on top of these features to perform object detection.
Since the diffusion model has been trained to generate semantically meaningful and class-aware bounding boxes, it provides valuable prior knowledge that substantially enhances object detection performance.

\subsubsection{Discussion of GenDet}
\begin{table*}[t]
	\footnotesize
	\centering
	\begin{tabular}{c|c|cccccc}
		\hline
		Type&Method & AP & AP$_{50}$ & AP$_{75}$ & AP$_{s}$ & AP$_{m}$ & AP$_{l}$\\
		\hline
		\multirow{2}*{Proposal-based}&Faster R-CNN~\citep{ren2016faster} & 40.2 &61.0 &43.8 &24.2 &43.5 &52.0 \\
		&Cascade R-CNN~\citep{cai2018cascade} & 44.3 &62.2 &48.0 &26.6 &47.7 &57.7 \\
		\midrule
		\multirow{2}*{Anchor-based}&RetinaNet~\citep{ross2017focal} & 38.7 & 58.0 & 41.5 & 23.3 & 42.3 & 50.3 \\
		&FreeAnchor~\citep{zhang2019freeanchor} & 38.7 & 57.3 & 41.5 & 21.0 & 42.0 & 51.3 \\
		\midrule
		\multirow{2}*{Center-based}&CenterNet~\citep{centernet} & 40.2 &58.3 &43.9 &23.4 &44.8 &51.6 \\
		&FCOS~\citep{fcos} & 42.3 &61.1 &45.4 &24.4 &45.9 &55.8 \\
		\midrule
		\multirow{3}*{Set-based}&DETR~\citep{detr} & 42.0 &62.4 &44.2 &20.5 &45.8 &61.1 \\
		&Sparse R-CNN~\citep{sun2021sparse} & 45.0 &63.4 &48.2 &26.9 &47.2 &59.5 \\
		&DiffusionDet~\citep{diffusiondet} & 45.8 &64.5 &50.8& 27.6& 48.7 &62.2 \\
		\midrule
		Generation-based &GenDet (Feature-based)&30.1 &45.6 &31.4 &10.2&34.5 &50.4 \\
		Set-based&GenDet (Learning-based)&46.4 &64.2 &50.5 &27.7&49.6 &63.2 \\
		\hline
	\end{tabular}
	\caption{Comparisons with different object detectors on
		COCO 2017 val set.}
	\label{eval_coco}
\end{table*}
Due to its slow detection speed, the reliance on complex post-processing, and its limited performance in challenging scenarios, GenDet cannot be considered a practical solution. Nevertheless, the primary contribution of our work lies in exploring the potential of generative models for object detection, specifically investigating how a single generative model can simultaneously perform both detection and generation tasks. We demonstrate the feasibility of applying Stable Diffusion to object detection, which may open a promising new research direction at the intersection of object detection and diffusion models.

\section{Experiments}

\subsection{Datasets}
GenDet is developed and evaluated on two widely used object detection benchmarks, COCO 2017~\citep{coco} and CrowdHuman~\citep{shao2018crowdhuman}.
\paragraph{COCO2017 dataset}
The COCO 2017 object detection dataset~\citep{coco} is a widely used benchmark that includes a total of 118,000 training images and 5,000 validation images. Each image in the dataset is annotated with bounding boxes that define the location of objects. On average, there are 7 objects per image, with some images containing as many as 63 objects. The dataset features objects of varying sizes, ranging from small to large, often within the same image. For evaluation, we report the Average Precision (AP) for bounding boxes, which is computed across multiple Intersection over Union (IoU) thresholds.
\begin{figure*}[t]
	\centering
	\includegraphics[width=0.9\linewidth]{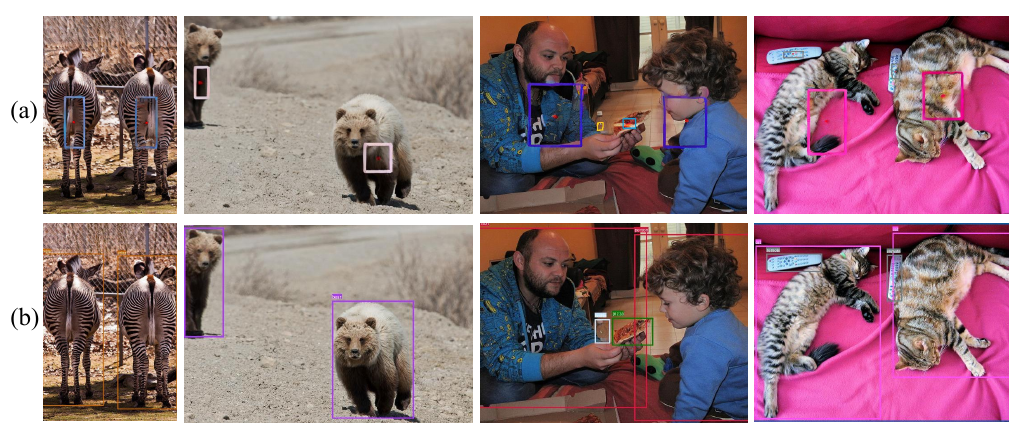}
	\caption{Visualization of object detection results using GenDet. Figure (a) displays the generated colored bounding boxes with scaled proportions, and Figure (b) shows the final object detection results, including object categories, after post-processing. As demonstrated by the results, GenDet effectively performs object detection tasks using the generative model. However, it does have some limitations, such as difficulty in detecting small objects.}
	\label{fig:overall1}
\end{figure*}
\paragraph{CrowdHuman dataset} 
The CrowdHuman dataset~\citep{shao2018crowdhuman} is a large-scale benchmark specifically designed for detecting pedestrians in crowded scenes. It contains over 15,000 training images and 4,300 validation images, with each image annotated with bounding boxes for full bodies, visible body parts, and head regions. A key challenge of this dataset lies in its high density of pedestrians, with an average of more than 20 persons per image and frequent heavy occlusions. These characteristics make it a widely adopted benchmark for evaluating pedestrian detection and occlusion handling methods.

\subsection{Implementation Details}
\label{Implementation}

We develop GenDet using Diffusers~\citep{diffusers} on COCO 2017~\citep{coco} and Crowdhuman~\citep{shao2018crowdhuman} object detection datasets, integrating Stable Diffusion v2~\citep{ldm} as the backbone while adhering to its original pre-training configuration with a \(v\)-objective~\citep{v}. Text conditioning is disabled to focus solely on visual inputs, and the training process strictly follows the methodology described in our approach. We employ multi-resolution noise strategies~\citep{marigold} to preserve low-frequency details. The DDPM noise scheduler~\citep{ddpm} with 1,000 diffusion steps is utilized during training, while inference leverages the DDIM scheduler~\citep{ddim}, reducing the sampling steps to 50 for faster computations.  
The training setup uses a batch size of 1, ensuring the input retains its original resolution throughout the process. Optimization is performed using the Adam optimizer with a learning rate of \(3 \times 10^{-5}\). To improve generalization, random horizontal flipping is applied as a data augmentation technique during training. 

\subsection{Experimental Results}
We compared GenDet with proposal-based, anchor-based, center-based, and set-based object detectors, as shown in Table~\ref{eval_coco}. 
\begin{table}[h]
	\footnotesize
	\centering
	\begin{tabular}{c|ccc}
		\hline
		Method & AP$_{50}$ & mMR & Recall\\
		\hline
		DETR~\citep{detr}&66.1 &80.6 &-  \\
		DiffusionDet~\citep{diffusiondet}&91.4 &45.7 &98.4 \\
		\midrule
		GenDet (Feature-based)& 52.3 &86.8 &70.1 \\
		\hline
	\end{tabular}
	\caption{Performance on CrowdHuman dataset.}
	\label{crowdhuman}
\end{table}

GenDet with learning-based post processing achieved superior performance, delivering the best overall object detection results. Compared to DiffusionDet~\citep{diffusiondet}, our method improves detection performance for large objects by 1.0 AP, demonstrating that GenDet has stronger context aggregation capabilities, which enables it to effectively locate large objects and, consequently, improve detection accuracy. Although GenDet with learning-based post-processing leverages a pre-trained diffusion model to learn the object distribution, it remains a discriminative approach, similar to DiffusionDet. We further analyze GenDet with feature-based post-processing, the first generation-based object detector. The results indicate that GenDet with feature-based post-processing achieves promising object detection performance, paving the way for further research into generation-based object detectors. Table~\ref{crowdhuman} presents the preliminary experimental results on the challenging Crowdhuman dataset~\citep{shao2018crowdhuman}, which is characterized by complex occlusion. The detection performance still requires further improvement.

Figure~\ref{fig:overall1} presents the colored bounding boxes generated by GenDet. This visual representation effectively showcases GenDet's ability to detect objects, validating that generative models can also perform object detection tasks traditionally handled by discriminative models. 
\begin{figure}[h]
	\centering
	\includegraphics[width=0.9\linewidth]{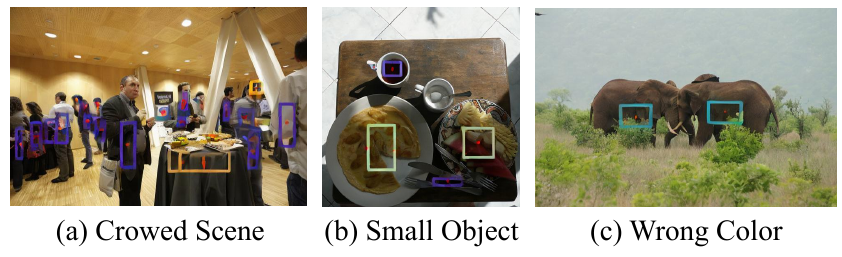}
	\caption{Illustrative examples of failure cases observed in GenDet with feature-based post-processing.}
	\label{fig:failure}
\end{figure}
Figure~\ref{fig:failure} illustrates several challenging scenarios that current models still struggle to handle effectively, such as object occlusion, crowded scenes, and small objects. Although the proposed learning-based post-processing can help detect objects in these cases, we leave the exploration of fully generative object detection approaches for future work.

\subsection{Ablation Studies}
\begin{table}[h]
	\footnotesize
	\centering
	\begin{tabular}{c|cccccc}
		\hline
		MGTO & AP & AP$_{50}$ & AP$_{75}$ & AP$_{s}$ & AP$_{m}$ & AP$_{l}$\\
		\hline
		&27.7 &44.1 &29.2  & 9.1 &31.0 &44.4 \\
		$\checkmark$& 28.7 &44.8 &30.2 & 9.4 &31.7 &45.4\\
		\hline
	\end{tabular}
	\caption{Ablation studies of multi-grained training objective (MGTO).}
	\label{abl_dual}
\end{table}
In this section, we present ablation experiments to quantitatively analyze the effectiveness of each component in GenDet. These experiments were conducted over 6 epochs.

\begin{table}[h]
	\footnotesize
	\centering
	\begin{tabular}{c|cccccc}
		\hline
		Ratio & AP & AP$_{50}$ & AP$_{75}$ & AP$_{s}$ & AP$_{m}$ & AP$_{l}$\\
		\hline
		$1/2$& 28.7 &45.1 &30.7 &9.6 &31.9 &45.0\\
		$1/3$& 29.0&45.6 &31.0 &9.8 &32.5 &45.7\\
		$1/4$& 28.5 &45.2 &30.9 &9.3 &32.1 &45.5\\
		\hline
	\end{tabular}
	\caption{Ablation study on the scaling ratio.}
	\label{abl_scal}
\end{table}
\paragraph{Multi-Grained Training Objective.}
Next, we examine the contribution of the multi-grained training objective. We compare the performance of the full model with a variant where the joint optimization of pixel-level reconstruction and detection-level semantic consistency is replaced by a single objective (\textit{e.g.}, pixel-level reconstruction only). The results in Table~\ref{abl_dual} indicate a drop in detection accuracy, underscoring the importance of jointly optimizing both generative and detection tasks.

\paragraph{Scaling Ratio.}

Lastly, we analyze the impact of scaling the detection box size, as shown in Table~\ref{abl_scal}. The results indicate that using a scaling ratio of one-third yields the best detection performance. A ratio of one-half is too large, leading to higher box overlap, while a ratio of one-quarter is too small, causing the boxes for small objects to be overly reduced and ultimately lowering detection accuracy.

\section{Conclusion}
In this paper, we introduced GenDet, a novel framework for object detection that harnesses the generative power of Stable Diffusion. It leverages a conditional generation architecture based on the pre-trained Stable Diffusion model, encoding detection tasks as semantic constraints. GenDet delivers precise manipulation of bounding boxes and object categories, improving both pixel-level accuracy and detection consistency. Experimental results demonstrate that GenDet surpasses traditional object detection methods, achieving state-of-the-art performance in object detection tasks.
This work paves the way for leveraging generative models in core computer vision tasks, highlighting the potential of generation-based frameworks in advancing object detection methodologies.

\paragraph{Limitations and future work.}
First, the detection speed of GenDet with feature-based post processing is relatively slow, requiring tens of seconds to process a single image, which limits its applicability in real-time scenarios.
Second, GenDet with feature-based post processing encounters difficulties in handling crowded scenes, occlusions, and small objects.
Third, the detection results exhibit some degree of randomness, which may affect consistency and reliability.
Fourth, the categories that GenDet with feature-based post-processing can detect are constrained by the color space.
In future work, we plan to address these challenges by optimizing inference efficiency, enhancing robustness under complex visual conditions, and reducing output variability, with the goal of further advancing generative object detection.

\bibliographystyle{IEEEtranN}
\bibliography{main}

\end{document}